\documentclass[journal]{IEEEtran}
\IEEEoverridecommandlockouts
\usepackage{amsmath,amssymb,amsfonts}
\usepackage{algorithmic}
\usepackage{graphicx}
\usepackage{makecell}
\usepackage{textcomp}
\usepackage{tikz}
\usepackage{textgreek}
\usepackage{pdflscape}
\usepackage{xcolor}
\usepackage[hidelinks]{hyperref}
\def\BibTeX{{\rm B\kern-.05em{\sc i\kern-.025em b}\kern-.08em
    T\kern-.1667em\lower.7ex\hbox{E}\kern-.125emX}}
    
\usepackage{multirow}
\usepackage{url}
\usepackage{color}
\usepackage[nolist,printonlyused]{acronym}
\usepackage{comment}
\usepackage{bm}
\usepackage{subcaption}
\usepackage{pgf}
\usepackage{lmodern}
\usepackage{import}
\usepackage[compress]{cite}

\usepackage{stfloats}
\usepackage{verbatim}
\begin{document}

\title{A Comparison of Neural Networks for \\ Wireless Channel Prediction\\ 
\thanks{\textit{The authors are with KTH Royal Institute of Technology. \\ Oscar Stenhammar and Gabor Fodor are also with Ericsson AB.}}
}

\author{
Oscar Stenhammar, Gabor Fodor and Carlo Fischione
}

\begin{acronym}
  \acro{2G}{Second Generation}
  \acro{3G}{3$^\text{rd}$~Generation}
  \acro{3GPP}{3$^\text{rd}$~Generation Partnership Project}
  \acro{4G}{4$^\text{th}$~Generation}
  \acro{5G}{5$^\text{th}$~Generation}
  \acro{AA}{Antenna Array}
  \acro{AC}{Admission Control}
  \acro{AD}{Attack-Decay}
  \acro{ADSL}{Asymmetric Digital Subscriber Line}
	\acro{AHW}{Alternate Hop-and-Wait}
  \acro{AMC}{Adaptive Modulation and Coding}
	\acro{AP}{Access Point}
  \acro{ANN}{artificial neural network}
  \acro{APA}{Adaptive Power Allocation}
  \acro{AR}{autoregressive}
  \acro{ARMA}{autoregressive moving average}
  \acro{ARIMA}{auto regressive integrated moving average}
  \acro{ATES}{Adaptive Throughput-based Efficiency-Satisfaction Trade-Off}
  \acro{AWGN}{additive white Gaussian noise}
  \acro{BPNN}{back propagation neural network}
  \acro{BB}{Branch and Bound}
  \acro{BD}{Block Diagonalization}
  \acro{BER}{bit error rate}
  \acro{BF}{Best Fit}
  \acro{BLER}{BLock Error Rate}
  \acro{BPC}{Binary power control}
  \acro{BPSK}{binary phase-shift keying}
  \acro{BPA}{Best \ac{PDPR} Algorithm}
  \acro{BRA}{Balanced Random Allocation}
  \acro{BS}{base station}
  \acro{CAP}{Combinatorial Allocation Problem}
  \acro{CAPEX}{Capital Expenditure}
  \acro{CBF}{Coordinated Beamforming}
  \acro{CBR}{Constant Bit Rate}
  \acro{CBS}{Class Based Scheduling}
  \acro{CC}{Congestion Control}
  \acro{CDF}{Cumulative Distribution Function}
  \acro{CDMA}{Code-Division Multiple Access}
  \acro{CL}{Closed Loop}
  \acro{CLPC}{Closed Loop Power Control}
  \acro{CNN}{convolutional neural network}
  \acro{CNR}{Channel-to-Noise Ratio}
  \acro{CPA}{Cellular Protection Algorithm}
  \acro{CPICH}{Common Pilot Channel}
  \acro{CoMP}{Coordinated Multi-Point}
  \acro{CQI}{Channel Quality Indicator}
  \acro{CRM}{Constrained Rate Maximization}
	\acro{CRN}{Cognitive Radio Network}
  \acro{CS}{Coordinated Scheduling}
  \acro{CSI}{channel state information}
  \acro{CSIR}{channel state information at the receiver}
  \acro{CSIT}{channel state information at the transmitter}
  \acro{CUE}{cellular user equipment}
  \acro{D2D}{device-to-device}
  \acro{DCA}{Dynamic Channel Allocation}
  \acro{DE}{Differential Evolution}
  \acro{DFT}{Discrete Fourier Transform}
  \acro{DIST}{Distance}
  \acro{DL}{downlink}
  \acro{DMA}{Double Moving Average}
	\acro{DMRS}{Demodulation Reference Signal}
  \acro{D2DM}{D2D Mode}
  \acro{DMS}{D2D Mode Selection}
  \acro{D-MIMO}{distributed multiple input multiple output}
  \acro{DPC}{Dirty Paper Coding}
  \acro{DRA}{Dynamic Resource Assignment}
  \acro{DSA}{Dynamic Spectrum Access}
  \acro{DSM}{Delay-based Satisfaction Maximization}
  \acro{ECC}{Electronic Communications Committee}
  \acro{EFLC}{Error Feedback Based Load Control}
  \acro{EI}{Efficiency Indicator}
  \acro{ELM}{extreme learning machine}
  \acro{eNB}{Evolved Node B}
  \acro{EPA}{Equal Power Allocation}
  \acro{EPC}{Evolved Packet Core}
  \acro{EPS}{Evolved Packet System}
  \acro{E-UTRAN}{Evolved Universal Terrestrial Radio Access Network}
  \acro{ES}{Exhaustive Search}
  \acro{FDD}{frequency division duplexing}
  \acro{FDM}{Frequency Division Multiplexing}
  \acro{FER}{Frame Erasure Rate}
  \acro{FF}{Fast Fading}
  \acro{FSB}{Fixed Switched Beamforming}
  \acro{FST}{Fixed SNR Target}
  \acro{FTP}{File Transfer Protocol}
  \acro{GA}{Genetic Algorithm}
  \acro{GBR}{Guaranteed Bit Rate}
  \acro{GLR}{Gain to Leakage Ratio}
  \acro{GOS}{Generated Orthogonal Sequence}
  \acro{GPL}{GNU General Public License}
  \acro{GRP}{Grouping}
  \acro{GRU}{gated recurrent unit}
  \acro{HARQ}{Hybrid Automatic Repeat Request}
  \acro{HMS}{Harmonic Mode Selection}
  \acro{HOL}{Head Of Line}
  \acro{HSDPA}{High-Speed Downlink Packet Access}
  \acro{HSPA}{High Speed Packet Access}
  \acro{HTTP}{HyperText Transfer Protocol}
  \acro{ICMP}{Internet Control Message Protocol}
  \acro{ICI}{Intercell Interference}
  \acro{ID}{Identification}
  \acro{IETF}{Internet Engineering Task Force}
  \acro{ILP}{Integer Linear Program}
  \acro{JRAPAP}{Joint RB Assignment and Power Allocation Problem}
  \acro{UID}{Unique Identification}
  \acro{IID}{Independent and Identically Distributed}
  \acro{IIR}{Infinite Impulse Response}
  \acro{ILP}{Integer Linear Problem}
  \acro{IMT}{International Mobile Telecommunications}
  \acro{INV}{Inverted Norm-based Grouping}
  \acro{IoT}{Internet of Things}
  \acro{IP}{Internet Protocol}
  \acro{IPv6}{Internet Protocol Version 6}
  \acro{ISD}{Inter-Site Distance}
  \acro{ISI}{Inter Symbol Interference}
  \acro{ITU}{International Telecommunication Union}
  \acro{JOAS}{Joint Opportunistic Assignment and Scheduling}
  \acro{JOS}{Joint Opportunistic Scheduling}
  \acro{JOELM}{jointly optimized extreme learning machine}
  \acro{JP}{Joint Processing}
  \acro{JS}{Jump-Stay}
  \acro{KF}{Kalman filter}
  \acro{KKT}{Karush-Kuhn-Tucker}
  \acro{L3}{Layer-3}
  \acro{LAC}{Link Admission Control}
  \acro{LA}{Link Adaptation}
  \acro{LC}{Load Control}
  \acro{LOS}{Line of Sight}
  \acro{LP}{Linear Programming}
  \acro{LS}{least squares}
  \acro{LSTM}{long short-term memory}
  \acro{LTE}{Long Term Evolution}
  \acro{LTE-A}{LTE-Advanced}
  \acro{LTE-Advanced}{Long Term Evolution Advanced}
  \acro{M2M}{Machine-to-Machine}
  \acro{MAC}{Medium Access Control}
  \acro{MANET}{Mobile Ad hoc Network}
  \acro{MC}{Modular Clock}
  \acro{MCS}{Modulation and Coding Scheme}
  \acro{MDB}{Measured Delay Based}
  \acro{MDI}{Minimum D2D Interference}
  \acro{MF}{Matched Filter}
  \acro{MG}{Maximum Gain}
  \acro{MH}{Multi-Hop}
  \acro{mMIMO}{massive multiple input multiple output}
  \acro{MIMO}{multiple input multiple output}
  \acro{MINLP}{Mixed Integer Nonlinear Programming}
  \acro{MIP}{Mixed Integer Programming}
  \acro{MISO}{Multiple Input Single Output}
  \acro{ML}{machine learning}
  \acro{MLP}{multilayer perceptron}
  \acro{MLWDF}{Modified Largest Weighted Delay First}
  \acro{MME}{Mobility Management Entity}
  \acro{MMSE}{minimum mean squared error}
  \acro{MOS}{Mean Opinion Score}
  \acro{MPF}{Multicarrier Proportional Fair}
  \acro{MRA}{Maximum Rate Allocation}
  \acro{MR}{Maximum Rate}
  \acro{MRC}{maximum ratio combining}
  \acro{MRT}{Maximum Ratio Transmission}
  \acro{MRUS}{Maximum Rate with User Satisfaction}
  \acro{MS}{mobile station}
  \acro{MSE}{mean squared error}
  \acro{MSI}{Multi-Stream Interference}
  \acro{MTC}{Machine-Type Communication}
  \acro{MTSI}{Multimedia Telephony Services over IMS}
  \acro{MTSM}{Modified Throughput-based Satisfaction Maximization}
  \acro{MU-MIMO}{multiuser multiple input multiple output}
  \acro{MU}{multi-user}
  \acro{NARX}{nonlinear autoregressive network with exogenous inputs}
  \acro{NAS}{Non-Access Stratum}
  \acro{NB}{Node B}
  \acro{NE}{Nash equilibrium}
  \acro{NCL}{Neighbor Cell List}
  \acro{NLP}{Nonlinear Programming}
  \acro{NLOS}{Non-Line of Sight}
  \acro{NMSE}{Normalized Mean Square Error}
  \acro{NN}{neural network}
  \acrodefplural{NN}[NNs]{neural networks}
  \acro{NORM}{Normalized Projection-based Grouping}
  \acro{NP}{Non-Polynomial Time}
  \acro{NRT}{Non-Real Time}
  \acro{NSPS}{National Security and Public Safety Services}
  \acro{O2I}{Outdoor to Indoor}
  \acro{OFDMA}{orthogonal frequency division multiple access}
  \acro{OFDM}{orthogonal frequency division multiplexing}
  \acro{OFPC}{Open Loop with Fractional Path Loss Compensation}
	\acro{O2I}{Outdoor-to-Indoor}
  \acro{OL}{Open Loop}
  \acro{OLPC}{Open-Loop Power Control}
  \acro{OL-PC}{Open-Loop Power Control}
  \acro{OPEX}{Operational Expenditure}
  \acro{ORB}{Orthogonal Random Beamforming}
  \acro{JO-PF}{Joint Opportunistic Proportional Fair}
  \acro{OSI}{Open Systems Interconnection}
  \acro{PAIR}{D2D Pair Gain-based Grouping}
  \acro{PAPR}{Peak-to-Average Power Ratio}
  \acro{P2P}{Peer-to-Peer}
  \acro{PC}{Power Control}
  \acro{PCI}{Physical Cell ID}
  \acro{PDF}{Probability Density Function}
  \acro{PDPR}{pilot-to-data power ratio}
  \acro{PER}{Packet Error Rate}
  \acro{PF}{Proportional Fair}
  \acro{P-GW}{Packet Data Network Gateway}
  \acro{PL}{Pathloss}
  \acro{PPR}{pilot power ratio}
  \acro{PRB}{physical resource block}
  \acro{PROJ}{Projection-based Grouping}
  \acro{ProSe}{Proximity Services}
  \acro{PS}{Packet Scheduling}
  \acro{PSAM}{pilot symbol assisted modulation}
  \acro{PSO}{Particle Swarm Optimization}
  \acro{PZF}{Projected Zero-Forcing}
  \acro{QAM}{Quadrature Amplitude Modulation}
  \acro{QoS}{Quality of Service}
  \acro{QPSK}{Quadri-Phase Shift Keying}
  \acro{RAISES}{Reallocation-based Assignment for Improved Spectral Efficiency and Satisfaction}
  \acro{RAN}{Radio Access Network}
  \acro{RA}{Resource Allocation}
  \acro{RAT}{Radio Access Technology}
  \acro{RATE}{Rate-based}
  \acro{RB}{resource block}
  \acro{RBG}{Resource Block Group}
  \acro{REF}{Reference Grouping}
  \acro{ReLU}{rectified linear unit}
  \acro{RLC}{Radio Link Control}
  \acro{RM}{Rate Maximization}
  \acro{RNC}{Radio Network Controller}
  \acro{RND}{Random Grouping}
  \acro{RNN}{recurrent neural network}
  \acro{RRA}{Radio Resource Allocation}
  \acro{RRM}{Radio Resource Management}
  \acro{RSCP}{Received Signal Code Power}
  \acro{RSRP}{Reference Signal Receive Power}
  \acro{RSRQ}{Reference Signal Receive Quality}
  \acro{RR}{Round Robin}
  \acro{RRC}{Radio Resource Control}
  \acro{RSSI}{Received Signal Strength Indicator}
  \acro{RT}{Real Time}
  \acro{RU}{Resource Unit}
  \acro{RUNE}{RUdimentary Network Emulator}
  \acro{RV}{Random Variable}
  \acro{SAC}{Session Admission Control}
  \acro{SCM}{Spatial Channel Model}
  \acro{SC-FDMA}{Single Carrier - Frequency Division Multiple Access}
  \acro{SD}{Soft Dropping}
  \acro{S-D}{Source-Destination}
  \acro{SDPC}{Soft Dropping Power Control}
  \acro{SDMA}{Space-Division Multiple Access}
  \acro{SE}{spectral efficiency}
  \acro{SER}{Symbol Error Rate}
  \acro{SES}{Simple Exponential Smoothing}
  \acro{S-GW}{Serving Gateway}
  \acro{SINR}{signal-to-interference-plus-noise ratio}
  \acro{SI}{Satisfaction Indicator}
  \acro{SIP}{Session Initiation Protocol}
  \acro{SISO}{single input single output}
  \acro{SIMO}{Single Input Multiple Output}
  \acro{SIR}{signal-to-interference ratio}
  \acro{SLNR}{Signal-to-Leakage-plus-Noise Ratio}
  \acro{SMA}{Simple Moving Average}
  \acro{SNR}{signal-to-noise ratio}
  \acro{SORA}{Satisfaction Oriented Resource Allocation}
  \acro{SORA-NRT}{Satisfaction-Oriented Resource Allocation for Non-Real Time Services}
  \acro{SORA-RT}{Satisfaction-Oriented Resource Allocation for Real Time Services}
  \acro{SPF}{Single-Carrier Proportional Fair}
  \acro{SRA}{Sequential Removal Algorithm}
  \acro{SRS}{Sounding Reference Signal}
  \acro{SU-MIMO}{single-user multiple input multiple output}
  \acro{SU}{Single-User}
  \acro{SVD}{Singular Value Decomposition}
  \acro{SVM}{support vector machine}
  \acro{SVR}{support vector machine for regression}
  \acro{TCP}{Transmission Control Protocol}
  \acro{TDD}{time division duplexing}
  \acro{TDMA}{Time Division Multiple Access}
  \acro{TDL}{tapped delay line}
  \acro{TETRA}{Terrestrial Trunked Radio}
  \acro{TP}{Transmit Power}
  \acro{TPC}{Transmit Power Control}
  \acro{TTI}{Transmission Time Interval}
  \acro{TTR}{Time-To-Rendezvous}
  \acro{TSM}{Throughput-based Satisfaction Maximization}
  \acro{TU}{Typical Urban}
  \acro{UE}{user equipment}
  \acro{UEPS}{Urgency and Efficiency-based Packet Scheduling}
  \acro{UL}{uplink}
  \acro{UMTS}{Universal Mobile Telecommunications System}
  \acro{URI}{Uniform Resource Identifier}
  \acro{URM}{Unconstrained Rate Maximization}
  \acro{UT}{user terminal}
  \acro{V2V}{vehicle-to-vehicle}
  \acro{V2X}{vehicle-to-everything}
  \acro{VR}{Virtual Resource}
  \acro{VoIP}{Voice over IP}
  \acro{WAN}{Wireless Access Network}
  \acro{WCDMA}{Wideband Code Division Multiple Access}
  \acro{WF}{Water-filling}
  \acro{WiMAX}{Worldwide Interoperability for Microwave Access}
  \acro{WINNER}{Wireless World Initiative New Radio}
  \acro{WLAN}{Wireless Local Area Network}
  \acro{WMPF}{Weighted Multicarrier Proportional Fair}
  \acro{WPF}{Weighted Proportional Fair}
  \acro{WSN}{Wireless Sensor Network}
  \acro{WWW}{World Wide Web}
  \acro{XIXO}{(Single or Multiple) Input (Single or Multiple) Output}
  \acro{ZF}{zero-forcing}
  \acro{ZMCSCG}{Zero Mean Circularly Symmetric Complex Gaussian}
\end{acronym}

\onecolumn
\noindent \LARGE \textbf{IEEE Copyright Notice}

\noindent \normalsize © 2023 IEEE. Personal use of this material is permitted. Permission from IEEE must be obtained for all other uses, in any current or future media, including reprinting/republishing this material for advertising or promotional purposes, creating new collective works, for resale or redistribution to servers or lists, or reuse of any copyrighted component of this work in other works

\thispagestyle{empty}
\clearpage
\setcounter{page}{1}
\twocolumn
\maketitle

\pagestyle{headings}
\markboth{IEEE WIRELESS COMMUNICATION MAGAZINE. PREPRINT VERSION. AUGUST 2023}%
{Shell \MakeLowercase{\textit{et al.}}: A Sample Article Using IEEEtran.cls for IEEE Journals}

\pagenumbering{arabic}

\begin{abstract}
The performance of modern wireless communications systems depends critically on the quality of the available channel state information (CSI) at the transmitter and receiver. Several previous works have proposed concepts and algorithms that help maintain high quality CSI even in the presence of high mobility and channel aging, such as temporal prediction schemes that employ neural networks. 
However, it is still unclear which neural network-based scheme provides the best performance in terms of prediction quality, training complexity and practical feasibility. To investigate such a question, this paper first provides an overview of state-of-the-art neural networks applicable to channel prediction and compares their performance in terms of prediction quality. 
Next, a new comparative analysis is proposed for four promising neural networks with different prediction horizons. The well-known tapped delay channel model recommended by the Third Generation Partnership Program is used for a standardized comparison among the neural networks.
Based on this comparative evaluation, the advantages and disadvantages of each neural network are discussed and guidelines for selecting the best-suited neural network in channel prediction applications are given.
\end{abstract}


\section*{Introduction}
As the sixth generation (6G) of wireless communication technologies and services emerges, higher expectations for mobile broadband services are set by end users.
Concepts of the evolving 5G and emerging 6G networks, such as distributed multiple input multiple output systems rely critically on the availability of up-to-date \ac{CSI}. However, obtaining accurate \ac{CSI} is non-trivial since the channel evolves over time as the scattering environment and the position of the \ac{UE} change. The evolution of the channel is often referred to as channel aging and poses a major challenge in the design of modern wireless systems.
To meet this challenge, channel prediction has emerged as a key tool to combat channel aging \cite{channelagingTruong, Bjorsell1647443}.
Most commonly, channel prediction is incorporated by exploiting time series of past channel estimations. Using predicted \ac{CSI}, it is possible to improve the performance of wireless communication, even in the presence of high mobility and rapidly changing channels. 
Accurately updated \ac{CSI} allows an adaptive transmitter to proactively tune the communication parameters, such as the transmit power, constellation size, and coding rate to enhance the network performance. 

A general and widely used method to characterize the evolution of the wireless channel is \ac{AR} models \cite{channelagingTruong}. This approach is model-based because it relies on analytical models of the dynamical evolution of the channel. In the modeled-based approach, the wireless channel is modeled as a linear combination of the previous realizations of the channel with some additive process noise. 
Conventionally, model-based methods such as Kalman and Wiener filtering have been used for channel prediction. 
Assuming that the channel evolves according to an \ac{AR} model with Gaussian noise and the second-order statistics of the \ac{AR} process are known or can be acquired, Kalman filtering is optimal in a \ac{MSE} sense \cite{4394065}. 
However, with increasing bandwidth and number of antennas, the complexity of Kalman filters grows relatively fast. Furthermore, the computational complexity is proportional to the square of the amount of previous channel data used in the model \cite{channelagingTruong}. 
To ensure satisfactory performance, it is often necessary to use higher-order models that employ many parameters. Thus, the inherent trade-off between the model order and the associated computational complexity often limits the performance of traditional model-based methods \cite{KvsML}. 

A recently proposed method to overcome channel aging is to use predictor antennas mounted on vehicles \cite{Bjorsell1647443}. The predictor antenna is specifically designed for vehicles moving at high speed and is typically placed on the exterior of the vehicle in front of the main antenna. 
In this way, the predictor antenna can estimate the channel from the position that the main antenna will reach soon. For vehicles moving at high speed, it is a valuable suggestion. However, mounting predictor antennas on legacy vehicles may not be viable in practice. Arguably, a more economical and viable solution is to find a prediction scheme with satisfactory performance and to update the software instead of installing hardware on existing connected vehicles.

In light of the above considerations, the increasing popularity and improvements of neural networks over the last years appear as a viable approach to wireless channel prediction. 
Specifically, by implementing neural networks according to the so-called data-driven approach, no underlying model needs to be assumed, as opposed to the model-based approach. This makes the predicting model less sensitive to disturbances and interference since it can learn from realistic data. In the case of predicting future channels based on solely the previous channel estimations, the channel prediction problem becomes a time series learning problem. Indeed, in the past years, channel prediction has been studied extensively and numerous techniques have been considered with the use of neural networks. For instance, channel prediction can also be conducted using the location of the \ac{UE}, which is appropriate for a static scenario.

However, in a real-world crowded urban environment, the spatial correlation of the channel can be very small or completely absent, due to moving objects. Also, by depending on the location of the \ac{UE}, an algorithm may become computationally more complex. By relying solely on historical time series data and the temporal correlation of the channel, channel prediction algorithms become computationally more efficient and scalable among different environments. For this reason, the present article aims to overview the most prominent neural networks methods and to identify research gaps in channel prediction that strictly uses historical channel measurements as input data. The most promising neural networks for channel prediction, which has performed well in previous studies, are compared using a dataset, with and without noise, simulated by the common and realistic \ac{3GPP} \ac{TDL}-A model \cite{tdla}. The performance of predicting fast-fading channels is studied over a large span of prediction horizons.

The advantages and disadvantages of each method are discussed to ultimately identify the most promising neural network for wireless channel prediction. This paper is, to the authors best knowledge, the first comparison of channel prediction methods that represent multiple different classes of neural networks. We provide a deeper understanding of the state-of-the-art in channel prediction to direct future research toward optimal models for real-world implementations.

To summarize, the contributions of this paper are the following:
\begin{itemize}
    \item An overview of previous works, focusing on channel prediction that employs data-driven \ac{ML} methods.
    \item An original quantitative comparison of the most promising data-driven methods identified from previous works. The data-driven methods are also compared to Kalman filtering.
    \item A discussion on how to develop the state-of-the-art in channel prediction, based on  numerical evaluation arguments. 
\end{itemize}

The rest of the sections in this paper are organized as follows: the representative state-of-the-art in channel prediction using \ac{ML} is overviewed; the prediction schemes that will be compared are described and justified; the proposed prediction schemes are numerically evaluated and compared by their performance; the outcome of the experiments are discussed; and finally, the results from the study we proposed in this paper are concluded.

\section*{Overview of previous works}

Model-based methods have been widely used to perform channel prediction. However, recent advances in \ac{ML} have accelerated several research areas, and recent studies have suggested that \ac{ML} has the potential to outperform conventional channel prediction model-based methods. 
The \ac{ML} model is a function that maps input data to an output decision or prediction, defined by its trainable parameters and its architecture. Training a supervised ML model means tuning the parameters to output a satisfactory output, usually by solving an optimization problem that minimizes a loss function.  In our case, the \ac{ML} model uses historical time series of channel measurements as input and outputs the future channel.

The performance of the model-based Kalman filter has been compared to a \ac{MLP} in \cite{KvsML}. The \ac{MLP} is a basic neural network that consists of several layers of nodes, where each node in one layer connects by a trainable parameter to every node in the following layer. To replicate the training process as with real channel data, \cite{KvsML} used noisy simulated data to train the \ac{MLP}. All other papers surveyed in our work that use simulated data assume perfect knowledge of the channel when training the neural network. The comparison of the Kalman filter and the \ac{MLP} method, with a small advantage to the Kalman filter, suggests a need for more advanced neural networks.

Several works have implemented more advanced structures for enhanced prediction accuracy. A popular model in image recognition is the \ac{CNN}, which in contrast to \ac{MLP} can take a matrix as input instead of a vector. It can learn to recognize patterns in smaller sections from an input matrix. By constructing a matrix of the size given by the time steps and the number of antennas, a \ac{CNN} is proposed in \cite{yuan2020machine} to predict \ac{AR} coefficients for channel evolution. 
Channel prediction has also been performed using a \ac{RNN} that utilizes the temporal correlation in sequential data, in contrast to the \ac{CNN}. A subset of frequency subcarriers was predicted individually by an \ac{RNN} in \cite{8746352}, followed by performing interpolation to predict the entire frequency domain used by the antenna.
Other works have combined \acp{CNN} and \acp{RNN} to predict the channel. Both \cite{luo2018channel, 8764345} have combined a \ac{CNN} with a \ac{LSTM} model, which is a type of \ac{RNN}. A well-known issue with \ac{RNN} is that it has training convergence issues due to vanishing or diverging gradients. \ac{LSTM} alleviates these problems. A comparison between \ac{LSTM} to conventional model-based methods has been proposed in \cite{liu2019deep}, where the effect of moving at different velocities were studied for channel prediction. In a recent paper \cite{9814839}, the authors modeled the channel between a \ac{UE} and \ac{BS} via a reconfigurable intelligent surface as a fast-fading channel using the \ac{LSTM}, assuming stationarity between the \ac{BS} and the reconfigurable intelligent surface. 
\begin{table*}[ht]
    \centering
    \caption{Contributions of the surveyed papers. Y means yes, N means no, S means simulated, M means measured.}
    \begin{tabular}{|c|c|c|c|c|c|c|c|}
    \hline
         Model & Performance & \makecell{Prediction \\ horizon} & \makecell{Noisy \\ label} & Mobility & \makecell{Data \\ generation} & \makecell{Prediction \\ procedure} & Paper \\ \hline \hline 

         LSTM & \makecell{Superior to ARIMA \\ and SVR} & 0.1-1ms & N & \makecell{Medium \\ \& High} & S & Time series & \cite{liu2019deep} \\ \hline

         Transformers & Superior to LSTM & \makecell{0.625-\\3.125ms} & N & Medium & S & Time series & \cite{9832933} \\ \hline 
         
         LSTM \& GRU & Superior to RNN & 1-5ms & N & High & S & Time series & \cite{jiang2020deep} \\ \hline
         
         RNN & Inferior to KF & 1.28ms & N & \makecell{Medium \\ \& High} & S & \makecell{Time series} & \cite{8746352} \\ \hline
         
         MLP & Inferior to KF & 40ms & Y & Low & S & Time series & \cite{KvsML} \\ \hline
         
         LSTM \& GRU & Superior to ARIMA & 1-10s & N & \makecell{Low \& \\ High} & S & \makecell{Encoder- \\decoder} & \cite{kulkarni2019deepchannel} \\ \hline 

         CNN/RNN & Superior to KF & - & N & \makecell{Low \& \\Medium} & S & Time series & \cite{yuan2020machine} \\ \hline
         
         LSTM/CNN & Superior to MLP & - & Y & None & M & Time series & \cite{luo2018channel} \\ \hline
         
         LSTM/CNN & \makecell{Inferior to LSTM \\ and CNN} & - & N & Medium & S & \makecell{UL-DL \& \\  subcarriers} & \cite{8764345} \\ \hline
         
         LSTM & \makecell{Superior to Minimum \\ Variance Unbiased} & - & N & Low & S & Time series & \cite{9814839} \\ \hline 
         
         C-GRBFnet & \makecell{Superior to LSTM} & - & N & None & S & \makecell{Spatial\\ prediction} & \cite{9791407} \\ \hline 

        \end{tabular}
    \label{tab_contributions}
\end{table*}
Aside from \ac{LSTM}, \ac{GRU} has also been proposed to improve the sequential \ac{RNN}, and is more computationally efficient than \ac{LSTM}. In channel prediction, the \ac{GRU} has been tested empirically by several researchers. The authors in \cite{kulkarni2019deepchannel} compared the \ac{LSTM} and \ac{GRU} with a proposed prediction model exploiting an encoder-decoder scheme, with \ac{LSTM} or \ac{GRU} layers at both the encoder and decoder side. Several datasets, one including 4G measurements, revealed a slight advantage to the \ac{LSTM}. In \cite{jiang2020deep}, an overview of channel prediction has been made where the \ac{LSTM} and \ac{GRU} have been studied over several prediction horizons. The first evaluation of a deep \ac{GRU} has been conducted, in favor of the \ac{GRU}. 

One additional type of neural network has been recently proposed in~\cite{9832933}, which adopts the transformer model to predict the channel. The transformer has the ability to predict multiple future time steps in parallel, by learning to identify and pay attention to critical behavior in sequential data. Another transformer-based model has been proposed in \cite{9791407} to predict the channel impulse response, based on the location of the \ac{UE}. It does not use historical channel measurements as input like the previously discussed papers but shows good results compared to the \ac{LSTM}.

In Table \ref{tab_contributions}, contributions from all papers are categorized in columns and summarized, with the topics of interest in this paper. As can be seen in the column \textit{Prediction procedure}, the channel is predicted using time series for all papers in Table \ref{tab_contributions}, except for \cite{9791407}. Although there are some papers using slightly different prediction procedures, time series are the foundation to make the prediction procedure more efficient. 

One conclusion from Table \ref{tab_contributions} is that the majority of the papers evaluate the channel prediction models based on simulated data. This is understandable since it is less costly and less time-consuming to collect simulated data. However, in a real-world implementation of channel prediction, measured data have to be used to conduct the predictions. The channel measurement and estimation process is unavoidably affected by noise. For this reason, noise was introduced in the training process in \cite{KvsML}, including the true data that is used to update the model. The column \textit{Noisy label} indicates whether the paper considers a noisy label for training the model. If the data is generated by measurements, the label is automatically noisy. There is only one paper that considers noisy labels while using a simulator to generate the dataset to train the prediction model. In our paper, we investigate its role and we show that it can have a major effect when evaluating the performance of the prediction methods.

The prediction horizons considered in Table \ref{tab_contributions} are almost exclusively correlated with mobility. If a paper considers high mobility of the \ac{UE}, the horizon is short, and vice versa, due to difficulties of predicting the channel over long horizons with a fluctuating channel. If the prediction horizon exceeds the coherence time, the channel's temporal correlation vanishes, and it becomes impossible to infer the channel out of current or past measurements. The prediction horizons of the papers listed in Table \ref{tab_contributions} are generally short. Half of the papers do not state on what time horizon the channel is predicted. Furthermore, no paper has included a prediction horizon long enough for the performance to fail. 

From the summary of previous works in Table \ref{tab_contributions}, research gaps can be found. First, it is not obvious which neural network is the most suitable for channel prediction. Second, although different data-driven models may each have good results, they have not been compared to each other. In most of the existing literature, data-driven models are compared to conventional model-based methods. The overviewed papers generally do not perform comparisons among data-driven models, or at best do partial comparisons. For example, \cite{jiang2020deep} compares \ac{LSTM} to a deep \ac{GRU}. The present paper is arguably the first to make a comprehensive comparison among the most prominent data-driven approaches.

\section*{Channel prediction using neural networks}
To identify the most promising neural network algorithm for the purpose of channel prediction, the most prominent algorithms found in the previous section are further analyzed and compared. Throughout the rest of this paper, five different regular types of neural networks are compared.

\begin{figure}[t]
    \centering
    \includegraphics[trim={9cm 4cm 6cm 5.5cm},clip,scale=0.38]{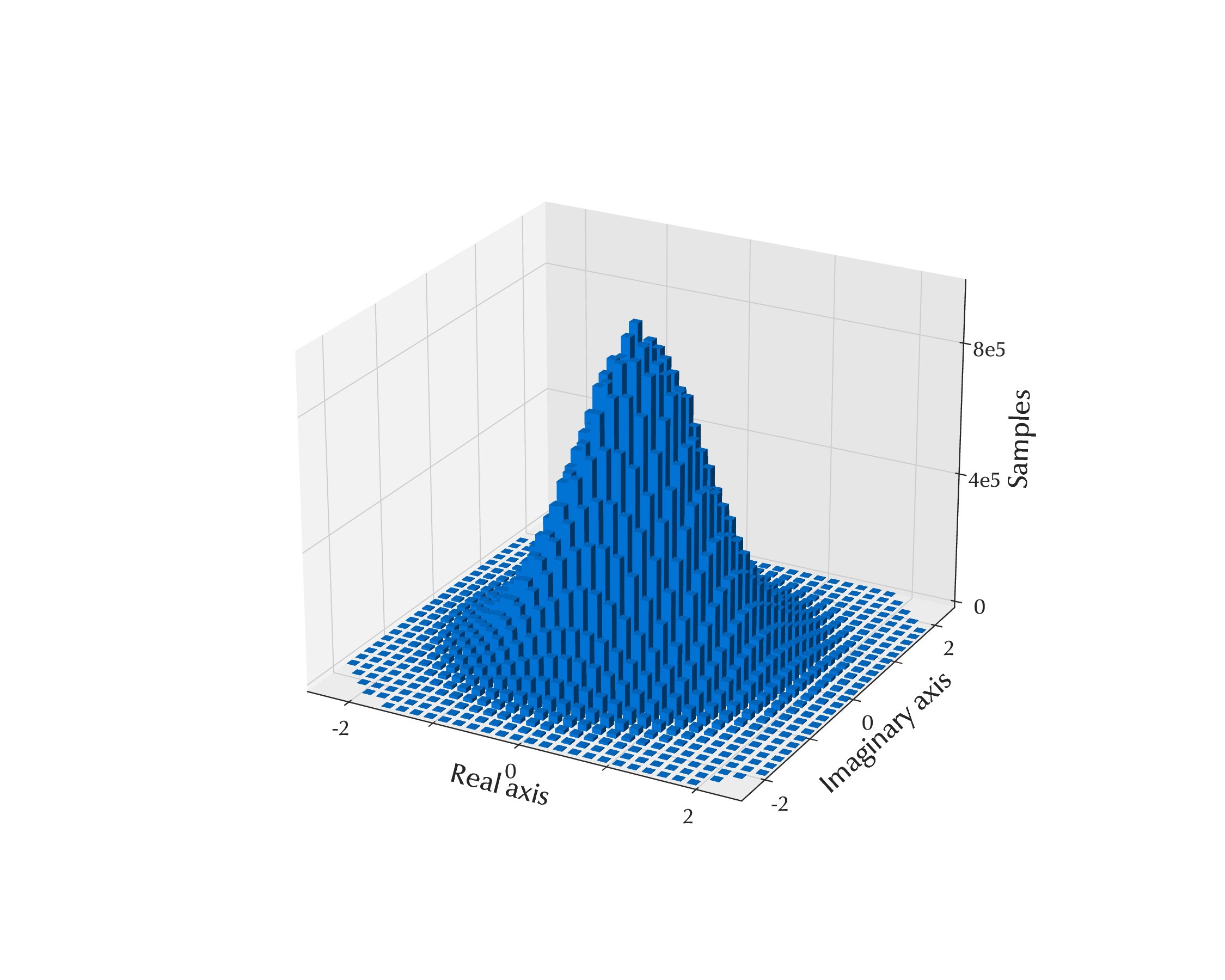}
    \caption{Distribution of simulated complex-valued channel. All samples in every subchannel and subcarrier are included in this distribution.}
    \label{H_distr}
\end{figure}

\begin{figure*}[t]
   \subfloat[\centering
   A sequence of the simulated channel from one antenna including the clean and distorted version.]{
	\begin{minipage}[0.8\width]{0.48\textwidth}
        \centering
        \includegraphics[width=1\textwidth]{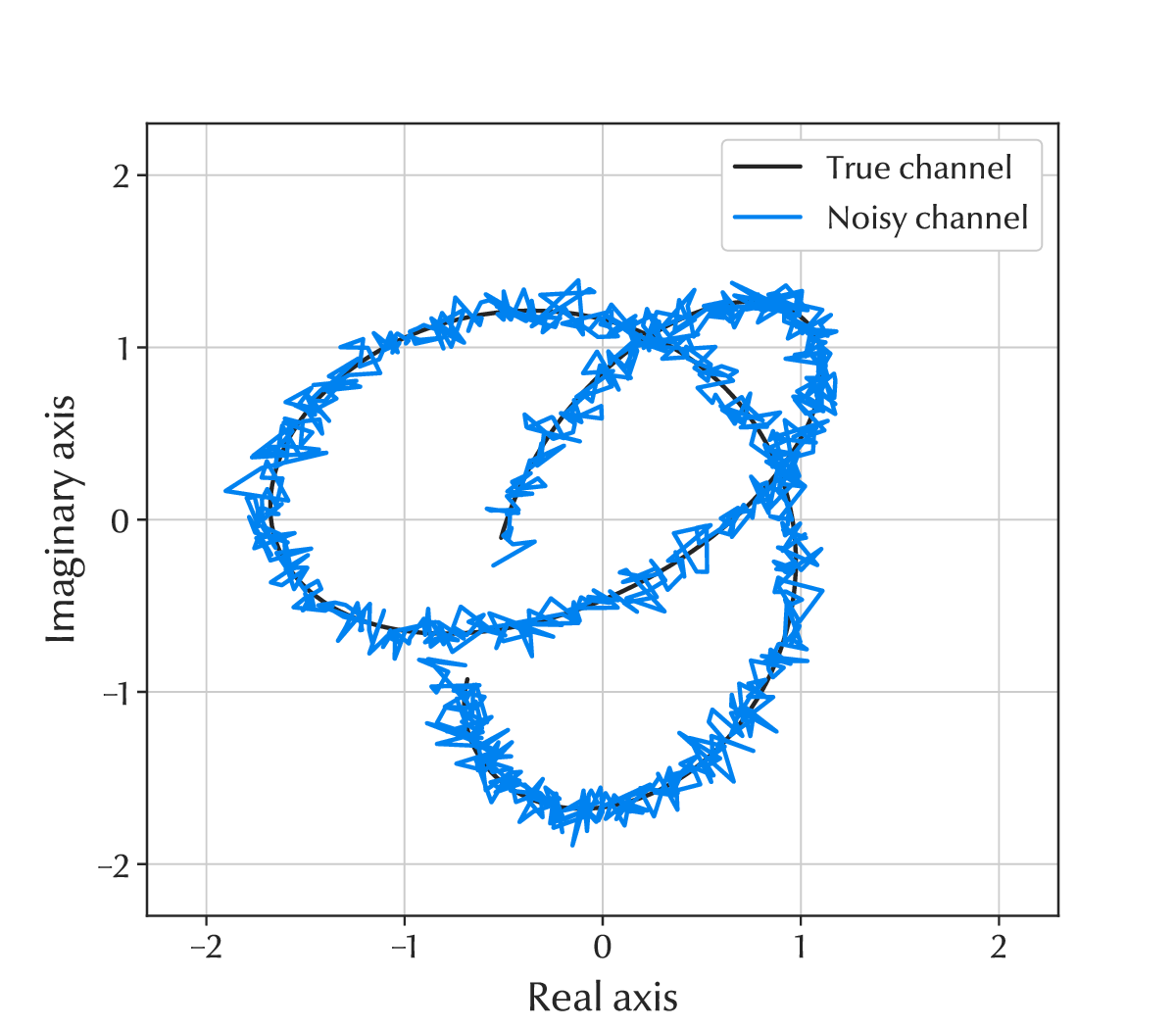}
        \label{Samples}
	\end{minipage}}
  \hfill
  \subfloat[\centering
  A sequence of the prediction of the noise-free channel for one antenna with the prediction horizon of 1 ms using \ac{GRU}.]{
	\begin{minipage}[0.8\width]{0.48\textwidth}
        \centering
        \includegraphics[width=1\textwidth]{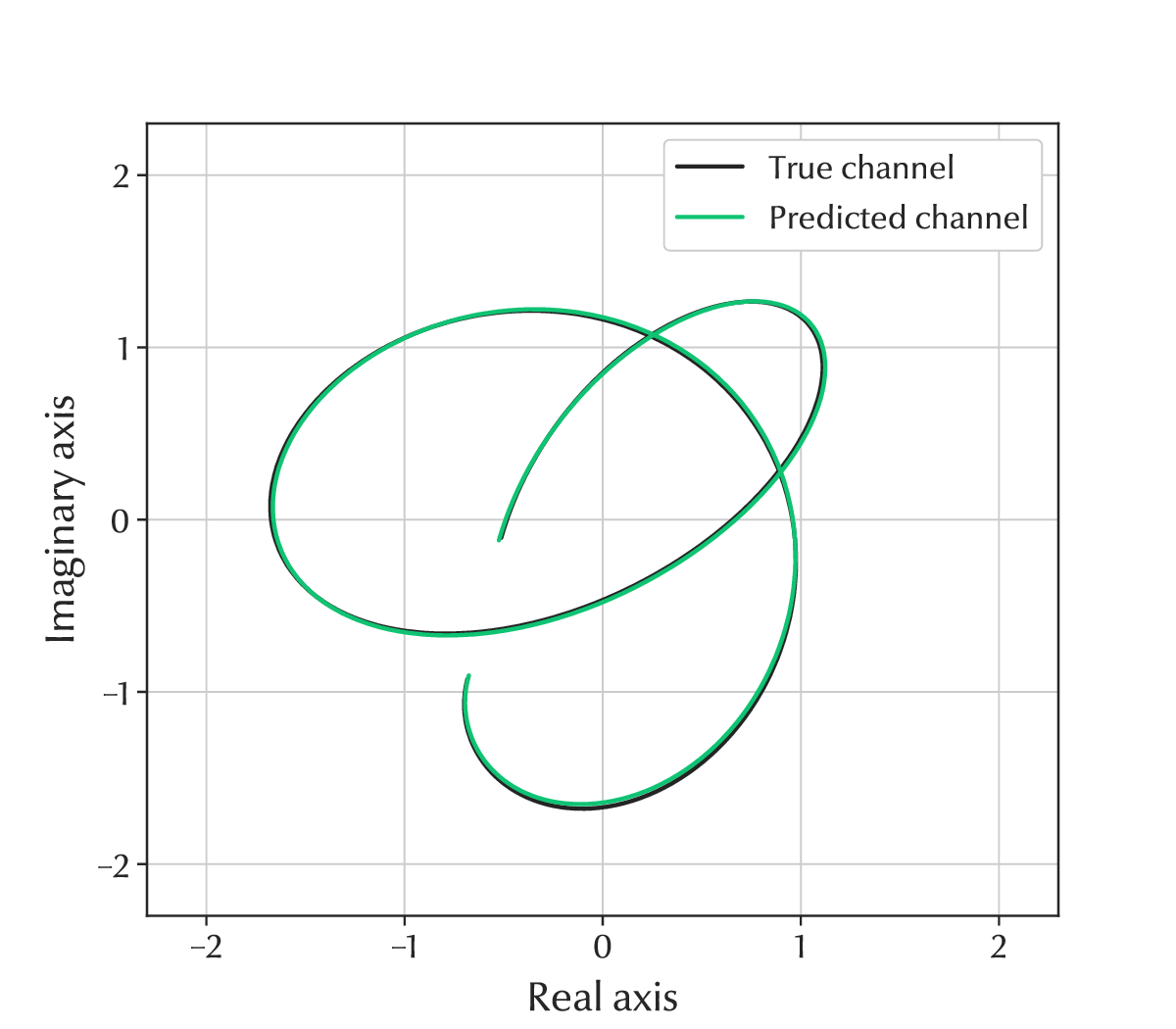}
        \label{tracking}
	\end{minipage}}
  \caption{The dynamics of the simulated channel and the predicted channel are visualized in this figure. Apart from the true channel in both sub-figures, one of the figures includes the channel distorted by additive Gaussian noise with \ac{SNR} of 20 dB, and the other includes the predicted channel. The duration of this particular sample is 53 ms.}
  \label{channel_tracking}
\end{figure*}

A common type of neural network is the feedforward neural network. It consists of a set of layers, each layer with multiple neurons, and basically constitutes the conceptual framework of all neural networks. All layers between the input- and output layers are called hidden layers. A feedforward neural network that has one or more hidden layers of neurons is called a \ac{MLP}. 
The \ac{MLP} has shown convincing results in many areas and is a very general framework compatible with many applications, since the input to a \ac{MLP} must be a one-dimensional vector. A vector that can be constructed from any type of data. However, the ordering of the elements in the input vector is disregarded. The result is that the potential importance of the position of the elements in the input vector is neglected.

One type of neural network that is built to preserve spatial data is the \ac{CNN}. The idea is to identify patterns in the input matrix, making the position of each element in the input matrix relevant compared to the \ac{MLP}. \acp{CNN} have acquired state-of-the-art status for their ability to detect patterns within image recognition. By using previous channel samples from multiple antennas, one can construct a matrix as a type of image as input to the \ac{CNN}. The advantages of the \ac{CNN} could be utilized to find patterns across time and among antennas.

A class of neural networks that is suitable for time series is the \ac{RNN} since it stores information from prior inputs in its internal state to influence the current output. That makes the \ac{RNN} able to benefit from sequential data better than the \ac{MLP} and \ac{CNN}. However, the classical \ac{RNN} has major problems with vanishing or exploding gradients. To solve these problems, modified networks have been suggested. One type of \ac{RNN}, that has reached state-of-the-art results in fields like speech recognition and language processing, is \ac{LSTM}. Every \ac{LSTM} cell is built by three gates, the input, output, and forget gate. As its name reveals, it has one long and one short-term memory. Since \ac{LSTM} was proposed, new algorithms with small modifications have been created. One of those, with promising results, is \ac{GRU}. The \ac{GRU} has two gates and one memory, making it computationally faster.

Another model that takes advantage of sequential data is the transfer model, a neural network architecture that is among the recently proposed prediction model~\cite{9832933}. It is most commonly used in natural language processing tasks but can also be applicable in time-series regression. It employs a self-attention mechanism to capture relationships among historical dynamics in a sequence. Multiple attention heads are used to capture different dependencies and relationships in parallel. The transformer incorporates positional encoding to convey the timeliness of each number. With its ability to capture long-range dependencies and parallel processing, the transformer has significant performance in time series prediction.

To evaluate and compare the performance of the promising neural networks for channel prediction, we consider a downlink MIMO scenario with $N_t$ antennas at the \ac{BS} and $N_r$ antennas at the \ac{UE}. We model the received signal as $\mathbf{y}[t] = \mathbf{H}[t]\mathbf{x}[t] + \mathbf{n}[t]$, 
where $\mathbf{x}[t]$ is the transmitted signal, $\mathbf{y}[t]$ is the received signal, $\mathbf{n}[t]$ is the additive noise and $\mathbf{H}[t]\in \mathbb{R} ^{N_t \times N_r} $ represents the channel. For various adaptive wireless technologies, outdated \ac{CSI} can cause heavy performance degradation. To obtain up-to-date \ac{CSI}, channel prediction is performed. To predict the future channel $\mathbf{\hat{H}}[t+p]$ on a desired prediction horizon $p$, we use $n$ historical measurements of the channel, indicated as $[\mathbf{H}[t-k \cdot n], ..., \mathbf{H}[t-k], \mathbf{H}[t]]$, where $k$ determines the time interval between each sample. Since the channel is complex-valued, real and imaginary values are separated in the input channel matrix $\mathbf{H}[t]$.

\section*{Experimental evaluation}
\label{results}
To evaluate the neural networks discussed in the previous section, fast-fading channel data is simulated using the standardized 3GPP \ac{TDL}-A model \cite{tdla}. The \ac{TDL}-A model is based on Rayleigh fading in a non-line-of-sight scenario and is useful when simulating the channel for cellular systems. The \ac{BS} and the \ac{UE} are assumed to have 2 antennas each, communicating at 2 GHz. The mobility of the \ac{UE} is set to 20 km/h, which gives a maximum Doppler shift of approximately 37 Hz. With 52 resource blocks, the number of subcarriers is 624.
To be consistent with the overviewed literature, only one subcarrier is considered at a time in the input and output of the prediction. This result is a dataset of around 26 milion data points. The distribution of the original dataset in the complex plane is plotted in Fig. \ref{H_distr}, showing the zero mean circularly symmetric complex Gaussian distribution of the channel. The histogram includes the samples from all MIMO channels and all subcarriers. The distribution symmetry and smoothness are results of the massive dataset. From this original dataset, the training dataset was randomly sampled to obtain 90000 samples, and the test dataset 10000 samples.

Experiments are conducted for two versions of the dataset, one using the original dataset and the other with the presence of noise. To imitate a realistic scenario and represent the uncertainties from channel estimation, the channel is distorted with Gaussian noise yielding an \ac{SNR} of 20 dB. The noise is present in the inputs and outputs of the training data. In this way, the model is trained with realistic noisy channel data, 
which to the authors knowledge, has not been studied earlier when using \acp{CNN}, \acp{LSTM} or \acp{GRU}. 
In the test data however, only the inputs are distorted to evaluate the predictor correctly. To reconstruct the predicted channel, the output vector from the neural networks is reshaped into a vector of complex-valued channels for each time instance. In Fig. \ref{Samples}, a small sample of the noisy-, true-, and predicted channels are plotted.

\begin{figure*}[t]
  \subfloat[\centering
  Prediction error of simulated noise-free channel data, as a function of the prediction horizon, for all considered models.]{
	\begin{minipage}[0.8\width]{0.47\textwidth}
        \centering
        \includegraphics[width=1\textwidth]{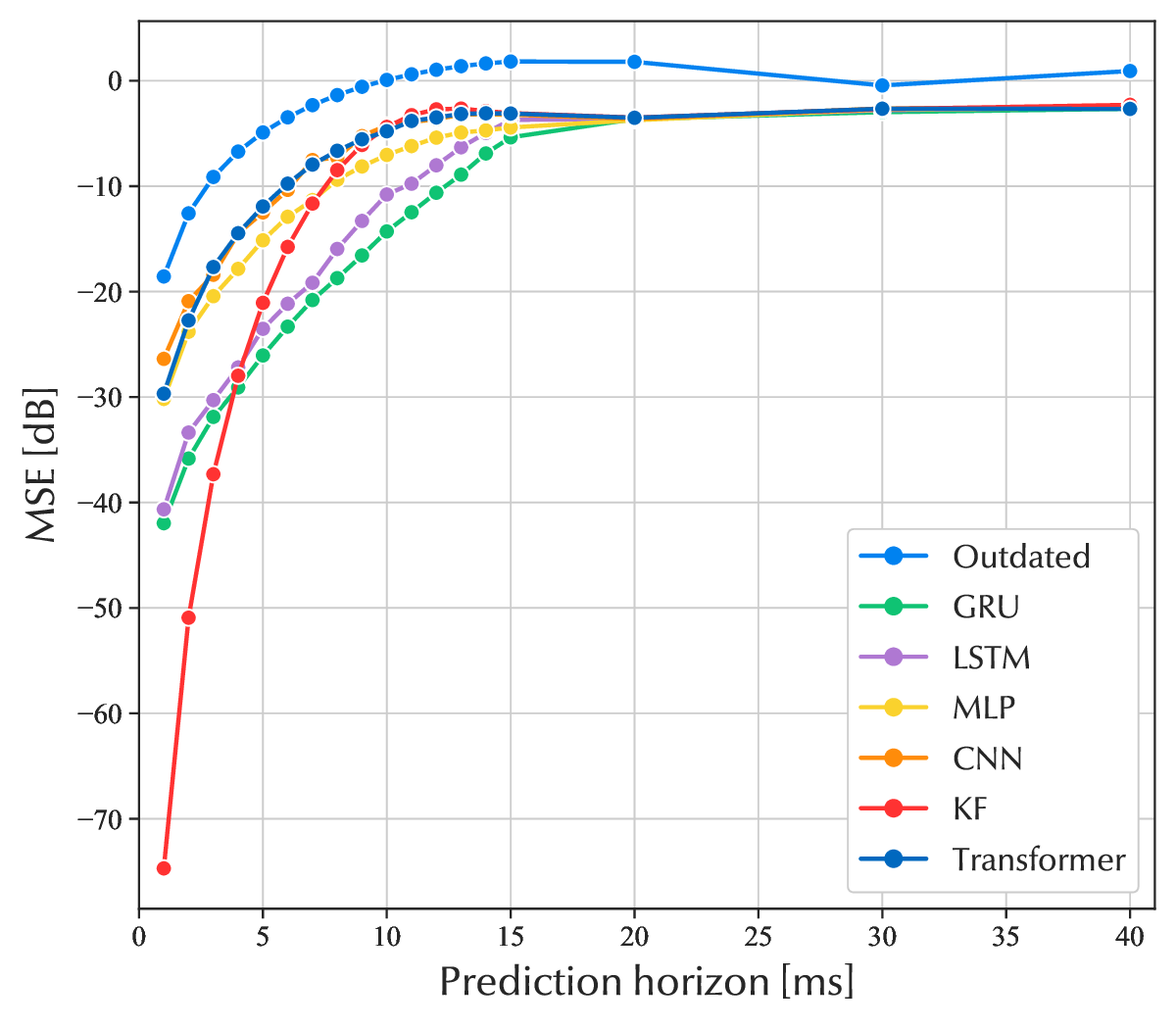}
        \label{horizon_clean}
	\end{minipage}}
 \hfill	
  \subfloat[\centering
  Prediction accuracy of the channel distorted by additive Gaussian noise with \ac{SNR} of 20 dB, as a function of the prediction horizon, for all considered  models.]{
	\begin{minipage}[0.8\width]{0.47\textwidth}
        \centering
        \includegraphics[width=1\textwidth]{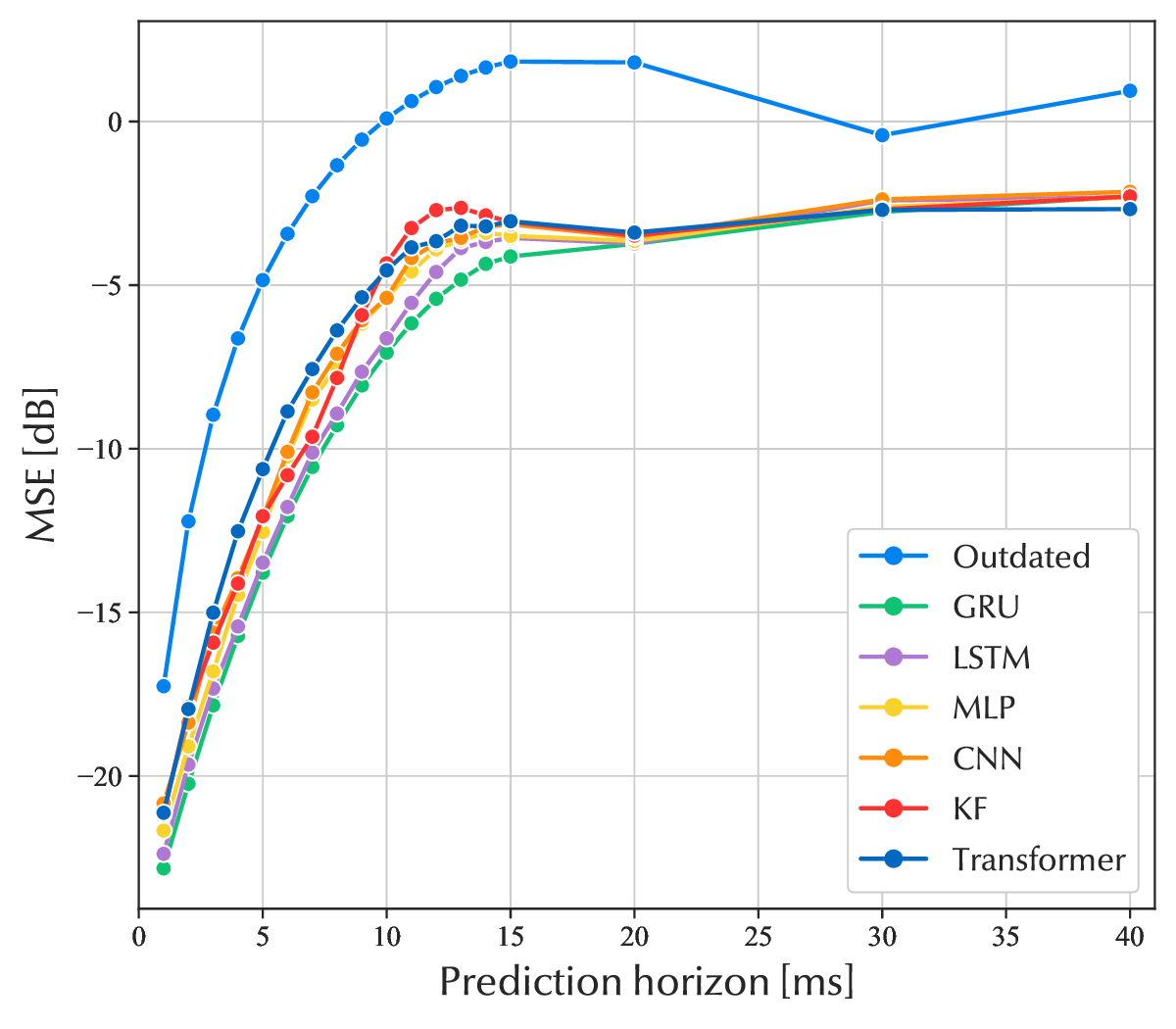}
        \label{horizon_snr}
	\end{minipage}}
\caption{The performance, measured in the \ac{MSE}, evaluated for all considered models and compared to each other, the Kalman Filter and the outdated channel.}
\label{performance}
\end{figure*}

With the Adam optimizer, the aim is to minimize the \ac{MSE} over 200 epochs. The number of historical channel measurements used to predict the future channel is $n=5$ with a sampling time of 1 ms throughout the paper. The neural networks are trained to optimize the accuracy for each prediction horizon. The \ac{MLP} is set to have 6 layers. For the \ac{CNN}, the number of convolutional layers is 4, each followed by a pooling layer, ending with 2 linear layers with \ac{ReLU} activation functions just as for the \ac{MLP}. The \acp{RNN} has 3 stacked layers. The hyperparameters of the models are tuned experimentally to obtain the best performance. For further insights of the models, the project can be found at Github\footnote{\url{https://github.com/osst3224/Channel_prediction_DNN.git}}. The number of hidden states is set to $150$ throughout this paper. The activation function used is the $tanh()$, as suggested in \cite{jiang2020deep}. Complexity analyses for the different models can be found in \cite{KvsML, 9832933,  jiang2020deep}. An empirical evaluation of the size of all models used in this paper is summarized in Table \ref{tab_complexity}.
The neural networks are trained for prediction horizons in the range of $1-40$ ms, with the purpose of investigating longer prediction horizons, which has been marginally done in previous works. 

\begin{table}[ht]
    \centering
    \caption{Empirical evaluation of the complexity for each prediction model.}
    \begin{tabular}{|c|c|c|}
    \hline
         Model & \makecell{Trainable \\ parameters} & \makecell{Elapsed time \\ per prediction} \\ \hline \hline
                           
         MLP & 1 990 402 & $27\mu s$  \\ \hline
         CNN & 4 338 & $33\mu s$  \\ \hline
         LSTM & 455 102 & $34\mu s$  \\ \hline 
         GRU & 341 402 & $28\mu s$  \\ \hline 
         Transformer & 463 874 & 97$\mu s$ \\ \hline
  
        \end{tabular}
    \label{tab_complexity}
\end{table}

The naive predictor, which assumes that the future channel is equal to the outdated (current) channel is used as a benchmark. To further evaluate the performance of neural networks in channel prediction, Kalman filtering is implemented as well.
For longer prediction horizons of several sampling times ahead, the channel prediction of 1 ms ahead is used as historical measurements for the next prediction. This procedure was repeated until the desired prediction horizon is reached, just as in \cite{KvsML}.
To fairly compare the performance of the Kalman filter and the neural networks, the \ac{MSE} of the Kalman filter was calculated after it had reached convergence. 

The \ac{MSE} for the test dataset is plotted in decibels as a function of the prediction horizon in Fig. \ref{horizon_clean}. The performance of the \acp{RNN} are similar, with a small advantage to the \ac{GRU}. The reason they are performing very similarly is that their architecture is closely related. With a little higher \ac{MSE}, the \ac{MLP}, \ac{CNN}, and transformer also have quite similar performance. The \ac{MSE} grows at a steady pace as the prediction horizon increases. 
As the prediction horizon grows, the channel's temporal correlation weakens, which naturally makes it more difficult to predict the channel. When the prediction horizon is around 15 ms, the performance of the neural networks reaches a level of error that stays relatively constant with a higher prediction horizon. This indicates that the coherence time of the channel is approximately 15 ms. The rate at which the channel ages is determined by the \ac{UE} mobility in this channel model. With a lower \ac{UE} mobility in the simulation, the performance of the neural networks would be better for longer prediction horizons.

The behavior is repeated for the performance of the neural networks trained with the noisy dataset, visualized in Fig. \ref{horizon_snr}. The performance of the neural networks is kept somewhat constant for prediction horizons longer than 15 ms. With shorter horizons, the performance of all neural networks is in the same relative order to each other in the noisy and noise-free case. The \ac{MSE} of the test data is substantially higher when noise is introduced in the training and test dataset. However, when noise is introduced and the prediction horizon is short, there is no significant difference in performance between the \acp{RNN}, \ac{MLP}, \ac{CNN}, and transformer. 

The Kalman filter performs very well in a noise-free environment on short prediction horizons. When the prediction horizon considered is over 5 ms, \ac{LSTM} and \ac{GRU} perform better than the Kalman filter. In a noisy environment, the Kalman filter behaves as the neural networks. But from Fig. \ref{performance}, it is apparent that the \acp{RNN} outperforms the Kalman filter over long prediction horizons.

\section*{Discussion}
\label{discussion}
From Fig. \ref{horizon_clean}, it is evident that the recurring memory cell gives a strong advantage in wireless channel time series prediction and constitutes a robust performance for the \ac{GRU} respectively \ac{LSTM} compared to the \ac{MLP}, \ac{CNN}, and transformer model, regardless of the prediction horizon. From these results, it is concluded that \ac{GRU} is the state-of-the-art in channel prediction. The intuitive explanation for this is the GRU’s innate ability to find correlations in sequential data. The GRU is custom-made to predict sequentially temporal data. It has fewer parameters than the similar \ac{LSTM} network, which makes the \ac{GRU} inclined to learn better and faster.
However, in form of practical feasibility in real base stations, due to constraints in computational power and energy consumption, it might be better to consider the \ac{MLP} out of the neural networks due to its low complexity, especially during training. The computational time and power spent to run the prediction model is crucial for real-world implementations. Future research could further extend the identified methods for real-world implementations such as quantization, continual learning and one-shot learning.

For the scenario with the distorted channel, the difference in error between the \ac{GRU} and the \ac{MLP} is constant around 1dB. The \acp{MLP} lower computational complexity might make it more suitable in a real-world implementation where computational power is limited due to time and energy constraints. As shown in \cite{KvsML}, the \ac{MLP} has lower computational complexity than the Kalman filter. Another fact that makes neural networks more suitable in a real-world implementation is that the channel might not always follow a smooth pattern as in Fig. \ref{channel_tracking}. Since the MLP is data-driven, it can learn easier than the Kalman filter in such a case.

This paper has considered the scenario of non-line-of-sight communication, for the standardized 3GPP TDL-A channel model. The simulated dataset consists of 26 million data points and is statistically sufficient to cover the scenario in TDL-A. Therefore, the results presented in this paper are general due to the generality of the TDL-A model. It would be beneficial if future research could investigate neural networks' robustness in wireless channel prediction, in the presence of abrupt changes in the communication environment due to the appearance or disappearance of dominant paths. In the case of appearing or disappearing dominant paths, the neural networks trained in this paper would only need five consecutive samples after the abrupt change, to reinitialize satisfactory predictions. The reason is that the input to the neural networks is of five samples. Moreover, line-of-sight communication is more static and fades slower than non-line-of-sight communication. Hence, neural network models trained in this paper will perform well in scenarios containing slower variations than those exhibited by the training dataset as well. The Kalman filter, on the other hand, requires more than five samples to converge to acceptable results, as seen in \cite{KvsML}.

\section*{Conclusions}
\label{conclusion}
This paper overviewed the most prominent research results from the literature on machine learning for channel prediction. The main advantage of machine learning is that it does not assume any underlying model, which makes it flexible and able to learn a model from the data itself. From simulations of the non-line-of-sight scenario of the 3GPP standardized TDL-A model, the neural networks were trained and validated. In the scenario of a noise-free channel, the numerical experiments of this paper established that two \ac{RNN}s, namely \ac{GRU} and \ac{LSTM}, achieved considerably better results for prediction horizons up to 15 ms than the \ac{MLP}, \ac{CNN}, and transformer model. However, the Kalman filter performs better than all neural networks up to the prediction horizon of 4 ms.
In the case of channel measurements with noisy data, the difference in performance between the neural networks was not as significant. However, from ordering the neural networks by their performance, the order was the same in the case with and without noise. The Kalman filter performed similarly as the \ac{MLP}, \ac{CNN}, and transformer model in the noisy case.

Ultimately, this overview suggests that the \ac{GRU} is most suitable to perform channel prediction and has the potential to be considered the most promising for future real-world implementations.
For future work, we plan to perform channel prediction using more than one carrier. Also, data-efficient machine learning models will have to be considered in future research if the predictions have to be made in resource-constrained wireless devices.

\section*{Acknowledgment}
This work was partially supported by the Wallenberg AI, Autonomous Systems and Software Program (WASP), funded by the Knut and Alice Wallenberg Foundation. The work of C. Fischione was also funded by Digital Futures KTH research center and SSF.

\bibliographystyle{ieeetr}
\bibliography{references}

\section*{Biographies}
\textsc{Oscar Stenhammar} [M] (oscar.stenhammar@ericsson
.com) is currently pursuing his Ph.D. at KTH Royal Institute of Technology as an industrial doctoral student, employed by Ericsson AB and affiliated with the Wallenberg AI, Autonomous Systems and Software Program (WASP). He received an M.S. in Engineering Physics at Uppsala University in 2021.

\textsc{Gabor Fodor} [SM] (gabor.fodor@ericsson.com) 
received a Ph.D. in electrical engineering from the Budapest University of Technology and Economics in 1998 and received a D.Sc. from the Hungarian Academy of Sciences in 2019. He is currently a Master Researcher with Ericsson Research and an Adjunct Professor with KTH Royal Institute of Technology, Stockholm, Sweden. He is currently serving as an Editor for IEEE Transactions on Wireless Communications and IEEE Wireless Communications. 

\textsc{Dr. Carlo Fischione} [SM] (carlofi@kth.se) is full Professor at KTH Royal Institute of Technology, Network and Systems Engineering, Stockholm, Sweden. He received a Ph.D. in Electrical and Information Engineering in 2005 from University of L’Aquila, Italy, and has held research positions at Massachusetts Institute of Technology, Cambridge, MA (2015); Harvard University, Cambridge, MA (2015); and University of California at Berkeley, CA (2004-2005 and 2007-2008). His research interests include applied optimization, wireless Internet of Things, and machine learning. He received the “2018 IEEE Communication Society S. O. Rice” award and the 2007 best paper award of IEEE Transactions on Industrial Informatics. 

\end{document}